%% file: main.tex
\documentclass[letterpaper, 10 pt, conference]{ieeeconf}  

\IEEEoverridecommandlockouts                              

\usepackage{graphics} 
\usepackage{graphicx}
\usepackage{tabularx}
\usepackage{booktabs}
\usepackage{colortbl}
\usepackage{epsfig} 
\usepackage{times} 
\usepackage{amsmath} 
\usepackage{mathtools,amssymb}  
\usepackage{subcaption}
\usepackage{svg}
\usepackage{url}
\usepackage{siunitx}
\usepackage{microtype}
\usepackage{flushend}
\usepackage{algorithm}
\usepackage{algpseudocode}
\usepackage{tikz}
\usetikzlibrary{shapes.geometric, arrows.meta, positioning}
\usepackage{hyperref}
\hypersetup{colorlinks=true,
	breaklinks=true,
	urlcolor= black,
	linkcolor= black,
	bookmarksopen=false,
	filecolor=black,
	citecolor=black,
	linkbordercolor=blue
}

\title{\LARGE \bf

CAD-Driven Co-Design for Flight-Ready Jet-Powered Humanoids
}

\author{Punith Reddy Vanteddu$^{1,2}$, Davide Gorbani$^{1,2}$, Giuseppe L'Erario$^{1}$,\\   Hosameldin Awadalla Omer Mohamed$^{1}$, Fabio Bergonti$^{1}$, Daniele Pucci$^{1,2}$ 
\thanks{$^{1}$Artificial and Mechanical Intelligence, Istituto Italiano di Tecnologia, Genoa, Italy {\tt\small firstname.surname@iit.it}}%
\thanks{$^{2}$School of Computer Science, University of Manchester, Manchester, UK
        }}

\begin{document}

\maketitle
\thispagestyle{empty}
\pagestyle{empty}

\begin{abstract}
This paper presents a CAD-driven co-design framework for optimizing jet-powered aerial humanoid robots to execute dynamically constrained trajectories. Starting from the iRonCub-Mk3 model, a Design of Experiments (DoE) approach is used to generate 5,000 geometrically varied and mechanically feasible designs by modifying limb dimensions, jet interface geometry (e.g., angle and offset), and overall mass distribution. Each model is constructed through CAD assemblies to ensure structural validity and compatibility with simulation tools. To reduce computational cost and enable parameter sensitivity analysis, the models are clustered using K-means, with representative centroids selected for evaluation. A minimum-jerk trajectory is used to assess flight performance, providing position and velocity references for a momentum-based linearized Model Predictive Control (MPC) strategy. A multi-objective optimization is then conducted using the NSGA-II algorithm, jointly exploring the space of design centroids and MPC gain parameters. The objectives are to minimize trajectory tracking error and mechanical energy expenditure. The framework outputs a set of flight-ready humanoid configurations with validated control parameters, offering a structured method for selecting and implementing feasible aerial humanoid designs. \looseness=-1

\end{abstract}
\section{Introduction}
\input{introduction}
\section{Background}
\input{background}
\section{Co-Design Framework}
\input{Co-design_framework}

\section{Results}
\input{Results}

\section{Conclusions}
\input{Conclusions}



\bibliographystyle{ieeetr}
\bibliography{references}


\end{document}

%% file: introduction.tex
The design of aerial humanoid robots presents distinct challenges compared to both conventional aerial vehicles and ground-based humanoids. In addition to tracking and stability requirements during flight, these systems must respect anthropomorphic geometry and accommodate complex actuation schemes such as jet propulsion. The placement of thrust interfaces and the distribution of mass directly affect flight dynamics, control authority, and mechanical feasibility. As such, the design and control of aerial humanoids are tightly coupled problems that require coordinated exploration of both structural and control spaces \cite{gupta2021embodied, sitti2021physical}.

The co-design of robotic systems—where both morphology and control are jointly optimized—has seen increasing interest across domains ranging from manipulators and grippers to legged and mobile robots. In many of these systems, morphology directly influences control performance, and co-optimization has been shown to improve task execution \cite{spielberg2017functional, sartore2023codesign, zhao2020robogrammar}. For example, in modular manipulators, task-specific morphology and base pose can be optimized simultaneously to improve manipulability and reduce joint effort \cite{10802089}. In robotic hands, co-evolution of mechanical structure and grasping behavior using reinforcement learning enhances adaptability across diverse tasks \cite{10631681}. In quadrupeds, introducing parallel compliance has been shown to increase motion robustness and explosiveness in tasks like pronking and hop-turns by coupling morphology with control planning~\cite{ding2025versatile}.

Despite these advances, most co-design frameworks focus on systems with limited dynamic complexity or low-dimensional design spaces. Applications to humanoid platforms remain sparse, particularly in aerial systems where mass distribution and thrust interface placement are tightly coupled with control dynamics. While upper-body co-design frameworks have been developed to optimize actuator configurations for motion retargeting \cite{10769829}, and mobile robot design has integrated perception, planning, and hardware constraints \cite{10930687}, these methods are not directly applicable to jet-powered humanoids. The added complexity of thrust-vectoring, underactuation, and flight dynamics introduces new constraints on design feasibility and evaluation.

\begin{figure}[t]
\centering
\includegraphics[trim=150 0 150 0, clip,width=0.48\textwidth]{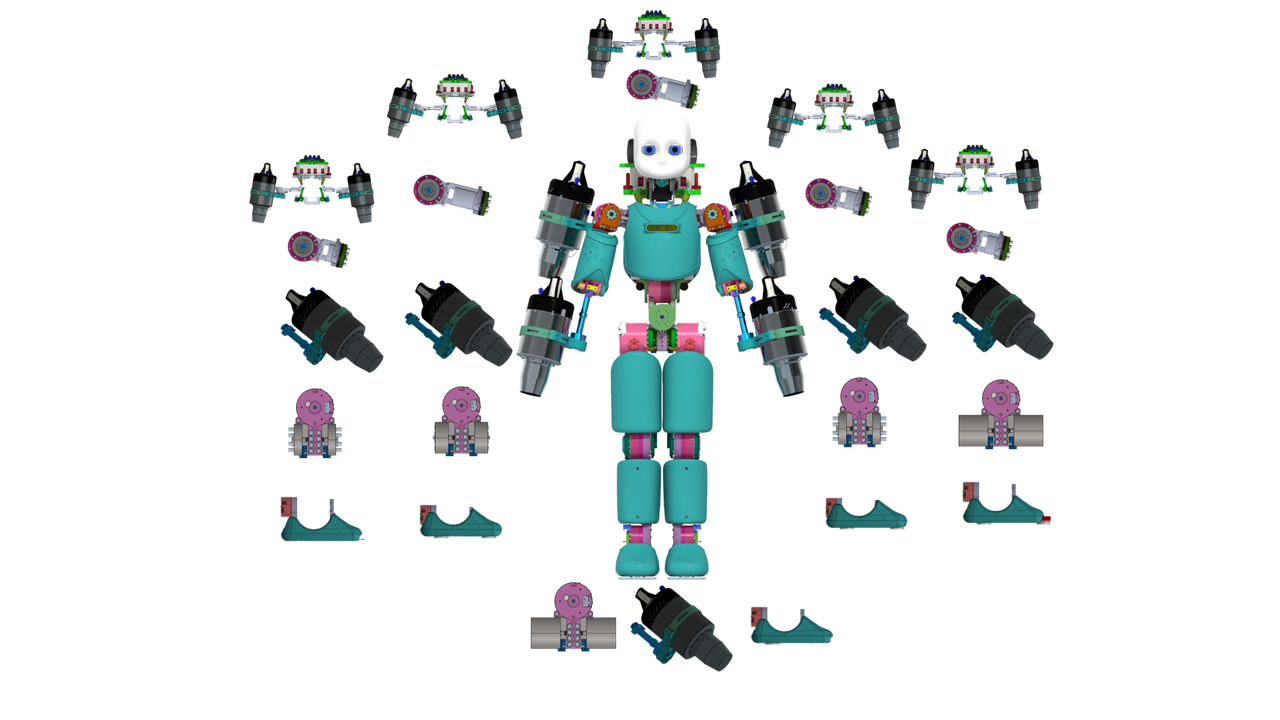}
\caption{Aerial humanoid with varied geometric assemblies obtained from a co-design framework.}
\label{Mk3-cad}
\end{figure}

Another challenge is the computational cost of exploring high-dimensional design spaces where each configuration must satisfy structural feasibility and dynamic requirements. Frameworks that use parallel training \cite{10631681}, differentiable simulation \cite{strgar2025accelerated}, or trajectory optimization \cite{10508097} address this issue in part but often rely on exploratory design loops that may not guarantee manufacturable outcomes. In contrast, recent work in legged robot compliance design \cite{10508097}, morphology-driven robustness \cite{ding2025versatile}, and robust multi-objective optimization \cite{chen2024multi, sathuluri2023robust} emphasizes solution feasibility and tolerance under uncertainty. However, these methods have not yet been extended to jet-powered humanoids, where morphology and trajectory performance must be optimized jointly under fabrication constraints.

Work has also been done to integrate CAD-informed geometry into co-design pipelines for jet-powered humanoids. These approaches focus on identifying control-relevant link parameters and filtering non-viable candidates using FEM-based structural analysis \cite{vanteddu2024cad}. Such pipelines validate the utility of CAD-informed modeling in co-design but primarily target component-level optimization and do not incorporate trajectory-level performance metrics across large-scale morphological variation.

In this work, we propose a framework that scales the design space and evaluates full-body CAD-derived configurations through flight-based performance metrics. Rather than optimizing isolated links, we vary whole-body geometry across thousands of models and assess each using a predictive controller in dynamic flight scenarios. The objective is to identify not only structurally feasible robots but also those capable of executing energy-efficient, dynamically constrained aerial trajectories.

We present a CAD-driven co-design framework for aerial humanoid robots that integrates parametric geometry generation, structured sampling via Design of Experiments (DoE)~\cite{montgomery2017design}, and trajectory-based evaluation using model predictive control. Starting from the iRonCub-Mk3 design \cite{pucci2017momentum, mohamed2021momentum}, we generate 5,000 physically feasible robot configurations by varying geometric parameters, such as link lengths and turbine placements, directly extracted from CAD. To reduce computational load, we apply K-means clustering~\cite{lloyd1982least} and select representative centroids. Each centroid is evaluated on a minimum-jerk trajectory using a linearized momentum-based MPC. A multi-objective optimization using NSGA-II \cite{deb2002fast} jointly searches for morphology and tunes control gain parameters to minimize trajectory tracking error and energy consumption. \looseness=-1

The remainder of this paper is organized as follows: Section II reviews background and related work. Section III details the proposed framework, including model generation, trajectory definition, control formulation, and optimization procedure. Section IV presents the results, and Section V concludes the paper with discussions and future directions.

%% file: background.tex
\begin{figure*}[ht]
      \centering
       \includegraphics[width=1\textwidth]{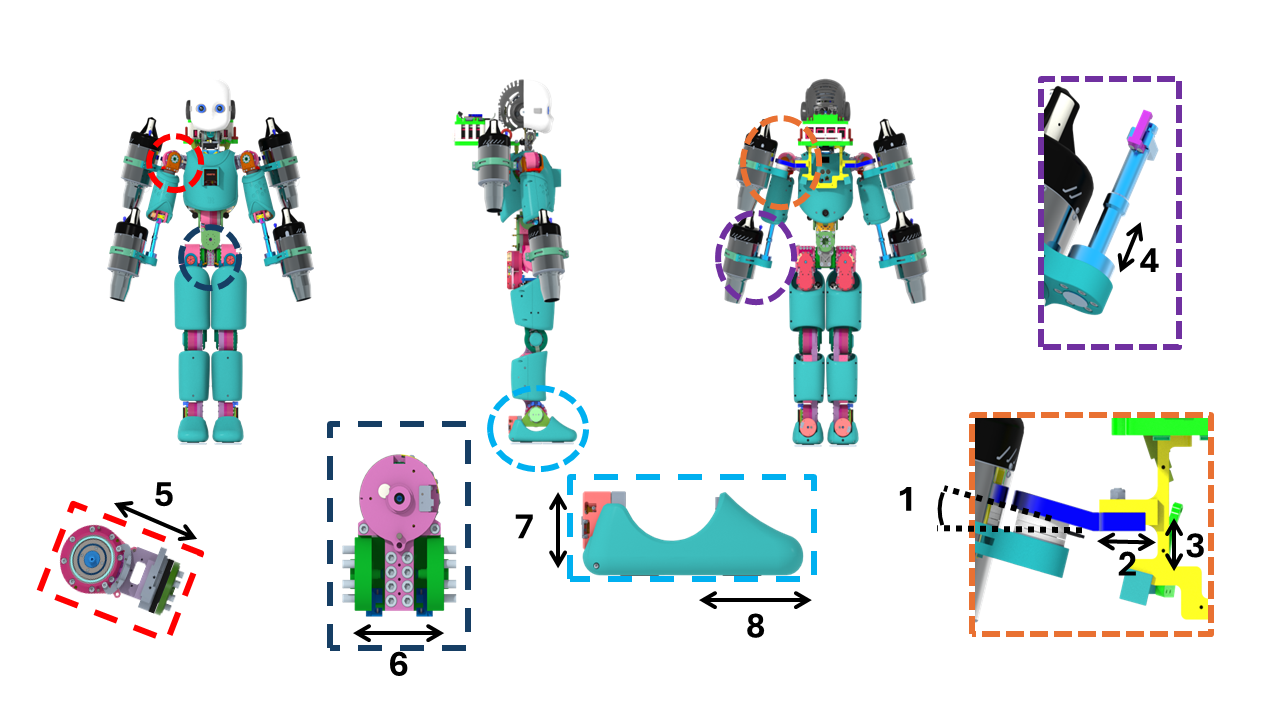}
      \caption{CAD assemblies of the links being modified. 1: Jetpack Turbine Angle; 2: Jetpack Turbine offset distance; 3: Jetpack Turbine height; 4: Forearm length; 5: Shoulder width; 6: Hip distance; 7: Ankle link height; 8: Foot length }
      \label{Mk3-cad}
   \end{figure*}
\subsection{Notation}
\begin{itemize}
    \item $S(x)$: Skew-symmetric matrix such that $x \times y = S(x)y$ for $x, y \in \mathbb{R}^3$.
    \item $\text{SO}(3)$: Special orthogonal group of 3D rotations.
    \item $\mathcal{I}$: Inertial frame; $\mathcal{B}$: Base frame attached to the robot.
    \item $\mathcal{G}[\mathcal{I}]$: Frame with origin at the CoM and orientation of $\mathcal{I}$.
    \item $R_{\mathcal{B}} \in \text{SO}(3)$: Rotation matrix from base frame to inertial frame.
    \item ${}^{\mathcal{G}[\mathcal{I}]}h$: Centroidal momentum, defined as $[{}^{\mathcal{G}[\mathcal{I}]}l; \, {}^{\mathcal{G}[\mathcal{I}]}w]$, where $l$ is linear and $w$ is angular momentum.
    \item ${}^{\mathcal{G}[\mathcal{B}]}l$, ${}^{\mathcal{G}[\mathcal{B}]}w$: Momentum terms expressed in body frame.
    \item $\mathbf{x}$: MPC state vector.
    \item $\mathbf{u}$: MPC input vector, composed of joint displacement $\Delta \mathbf{s}$ and turbine throttle $u_{th}$.
\end{itemize}

\subsection{Robot Model and Flight Control}

The robot used in this study is based on the iRonCub-Mk3 aerial humanoid, modeled as a floating-base multibody system of $n+1$ rigid links and $n$ actuated joints. Turbine thrusters are rigidly mounted on specific links and provide thrust forces during flight. The configuration vector is defined as $q := (p_B, R_B, s)$, where $p_B \in \mathbb{R}^3$ and $R_B \in \text{SO}(3)$ define the base position and orientation, and $s \in \mathbb{R}^n$ the joint angles. The velocity vector is $\nu := (\dot{p}_B, \omega_B, \dot{s})$.

The system dynamics are formulated in terms of centroidal momentum, capturing the influence of all external forces. The total momentum is given by:
\[
{}^{\mathcal{G}[\mathcal{I}]} h = \begin{bmatrix}
{}^{\mathcal{G}[\mathcal{I}]} l \\
{}^{\mathcal{G}[\mathcal{I}]} w
\end{bmatrix}, \quad
{}^{\mathcal{G}[\mathcal{I}]} \dot{h} = 
\begin{bmatrix}
mge_3 + \sum_{k=1}^{n_p} F_k \\
\sum_{k=1}^{n_p} S(r_k) F_k
\end{bmatrix}
\]
where \( g \) is the acceleration due to gravity and \( e_3 = \begin{bmatrix} 0 & 0 & 1 \end{bmatrix}^\top \) is the unit vector along the z-axis. $F_k = {}^{\mathcal{I}} a_k(q) T_k$ is the $k$-th thrust force  composed of its direction ${}^{\mathcal{I}}a_{k} \in \mathbb{R}^{3\times3}$ and the thrust intensity $T_{k} \in \mathbb{R}^+$. and $S(r_k)$ is the torque arm.

For trajectory tracking, a linearized Model Predictive Control (MPC) scheme is implemented. The state vector includes:
\[
\begin{aligned}
\mathbf{x} &= \begin{bmatrix}
{}^{\mathcal{W}}\mathbf{x}_{com} &
{}^{\mathcal{G}[\mathcal{I}]}\mathbf{l} &
\boldsymbol{\phi} &
{}^{\mathcal{G}[\mathcal{I}]}\boldsymbol{\omega} &
T &
\dot{T} &
e_{x} &
e_{\phi} &
\alpha_g
\end{bmatrix}^{\top}, \\
\mathbf{u} &= \begin{bmatrix}
\Delta \mathbf{s} &
u_{th}
\end{bmatrix}^{\top}
\end{aligned}
\]
\begin{equation}
    \begin{cases}
        \dot{e}_{x} = {}^{\mathcal{W}}\mathbf{x}_{com} - {}^{\mathcal{W}}\mathbf{x}_{com,ref} \\
        \dot{e}_{\phi} = \boldsymbol{\phi} - \boldsymbol{\phi}_{ref}
    \end{cases}
\end{equation}
where $\dot{e}_x$ is the CoM position error in the world frame, and $\dot{e}{\phi}$ is the orientation error expressed in terms of roll, pitch, and yaw deviation from the reference trajectory.
The controller minimizes a quadratic cost:
\begin{equation}
\begin{aligned}
J(\mathbf{x}, \mathbf{u}) =
& \frac{1}{2} \left\| {}^{\mathcal{W}}\mathbf{x}_{com} - {}^{\mathcal{W}}\mathbf{x}_{com,ref} \right\|^2_{\mathcal{W}_x}
+ \frac{1}{2} \left\| {}^{\mathcal{G}[\mathcal{I}]}\mathbf{l} \right\|^2_{\mathcal{W}_l} \\
& + \frac{1}{2} \left\| \boldsymbol{\phi} - \boldsymbol{\phi}_{ref} \right\|^2_{\mathcal{W}_\phi}
+ \frac{1}{2} \left\| {}^{\mathcal{G}[\mathcal{I}]}\boldsymbol{\omega} \right\|^2_{\mathcal{W}_\omega} \\
& + \frac{1}{2} \left\| \Delta \mathbf{s} \right\|^2_{\mathcal{W}_{\Delta s}}
+ \frac{1}{2} \left\| u_{th,k+1} - u_{th,k} \right\|^2_{\mathcal{W}_{u_{th}}} \\
& + \|e_{x} \|^{2}_{W_{e_{x}}} + \|e_{\phi} \|^{2}_{W_{e_{\phi}}}
\end{aligned}
\end{equation}

This formulation enables trajectory tracking while accounting for turbine dynamics and physical constraints. The tuning of the cost weights $ \mathcal{W}_x, \mathcal{W}_l, \ldots$ plays a key role in shaping the flight response.

\subsection{Geometric Parameters and Control Weights}

The robot's geometry is defined through a set of parametric variables derived from the iRonCub CAD model. These parameters influence link dimensions, thrust placement, and inertia distribution. The selected geometric input parameters are:\looseness=-1

\begin{itemize}
    \item Jetpack Turbine angle
    \item Jetpack Turbine offset distance
    \item Jetpack Turbine height
    \item Forearm length
    \item Shoulder width
    \item Hip distance
    \item Ankle link height
    \item Foot length
\end{itemize}

These parameters are systematically varied to generate a set of robot configurations, where each variation induces changes in the robot’s mass distribution, inertia, and turbine placement. For every model, the updated inertial properties are extracted automatically from CAD, including mass, center of mass, and link inertia. All models have been validated using Finite Element Method (FEM) analysis to ensure structural integrity and manufacturability at the assembly level. This guarantees that the solutions explored in the subsequent co-design process are not only dynamically diverse but also mechanically feasible for prototyping.

In addition to morphological variation, the MPC controller includes tunable weight matrices that penalize tracking errors and control effort. These weights define how the controller prioritizes state regulation versus input smoothness across the prediction horizon. Specifically, the tunable weights include:

\begin{itemize}
    \item \textbf{State error weights:} $\mathcal{W}_x$ (CoM position error), $\mathcal{W}_l$ (linear momentum error), $\mathcal{W}_\phi$ (orientation error), $\mathcal{W}_\omega$ (angular momentum error), $\mathcal{W}_{e_x}$ (integral of position error), $\mathcal{W}_{e_\phi}$ (integral of orientation error)

    \item \textbf{Input weights:} $\mathcal{W}_{\Delta s}$ (joint motion), $\mathcal{W}_{u_{th}}$ (throttle rate change)
\end{itemize}

These weights are optimized alongside model selection to balance tracking accuracy and energy consumption for each centroid configuration.

Together, these define the 16-dimensional input space of the co-design framework, composed of 8 geometric and 8 control parameters. In the following section, we describe how these inputs are jointly optimized using trajectory performance metrics.

%% file: Co-design_framework.tex
This section describes the pipeline used to jointly evaluate robot morphology and control performance. The co-design framework integrates three main components: (i) generation of structurally valid robot models, (ii) definition of a reference trajectory, and (iii) a multi-objective optimization that selects optimal design–control pairs based on performance metrics.

The design space for aerial humanoid robots is constructed to explore how morphological variations affect flight performance under jet propulsion. Starting from the iRonCub-Mk3 platform as a baseline, we define a set of eight geometric parameters related to turbine interface placement and limb proportions. These parameters were selected based on their influence on thrust orientation, mass distribution, and moment of inertia.

\subsection{Geometric Variation and Model Sampling}
\label{sec:geo_sampling}

To systematically explore the configuration space, we vary the eight input geometric parameters listed in Table \ref{geom_param}. These include three parameters related to jetpack placement and five related to limb geometry. Sampling is performed using Uniform Latin Hypercube (ULH) \cite{mckay2000comparison} to ensure that combinations of parameters are well-distributed across the multidimensional space. This process yields 5,000 unique design instances.

\begin{table}[t]
    \centering
    \caption{Geometric Parameters and Sampling Intervals}
    \label{geom_param}
    \begin{tabular}{l|c|c|c}
        \toprule
        \textbf{Parameter} & \textbf{Min} & \textbf{Max} & \textbf{Step} \\
        \midrule
        \rowcolor{gray!15}
        Jetpack Turbine angle [\qty{}\degree] & 0 & 14 & 2 \\
        Jetpack Turbine offset distance [\qty{}\mm] & 0 & 30 & 5 \\
        \rowcolor{gray!15}
        Jetpack Turbine height [\qty{}\mm] & 0 & 30 & 5 \\
        Forearm length [\qty{}\mm] & 0 & 40 & 5 \\
        \rowcolor{gray!15}
        Shoulder width [\qty{}\mm] & 0 & 50 & 5 \\
        Hip distance [\qty{}\mm] & 0 & 50 & 5 \\
        \rowcolor{gray!15}
        Ankle link height [\qty{}\mm] & 0 & 50 & 5 \\
        Foot length [\qty{}\mm] & 0 & 100 & 10 \\
        \bottomrule
    \end{tabular}
\end{table}

For each of the 5,000 generated configurations,  updated physical properties (Mass, Inertia) are directly derived from the CAD assemblies and automatically propagated to URDF \cite{ros_urdf} and MuJoCo XML files \cite{todorov2012mujoco} for compatibility with kinematic and physics-based simulation environments. In addition to inertial accuracy, we apply Finite Element Method (FEM) validation at the assembly level to ensure that each model satisfies structural feasibility under nominal loading. Models failing structural or assembly constraints are discarded.

To streamline the co-design optimization, each model is assigned a unique index. Rather than optimizing geometry directly, the algorithm selects from this validated pool of models by index, ensuring that only feasible configurations are considered during control and performance evaluation.

\subsection{Design Space Reduction via Geometric Clustering}
\label{sec:model_reduction}

Although all 5,000 robot variants are structurally feasible and simulation-ready, evaluating every model in the optimization pipeline would be computationally intractable. To address this, we apply model reduction using K-means clustering over the eight-dimensional geometric parameter space. Each robot is represented by a feature vector of normalized geometric inputs, and the clustering is performed to identify a set of representative configurations—or centroids—that span the diversity of the original dataset.

The centroids are used in place of the full model set during multi-objective optimization, enabling a tractable yet representative evaluation of the design space. By reducing the number of configurations while preserving the geometric diversity, this approach balances exploration coverage with computational efficiency.

In addition to model reduction, clustering supports sensitivity analysis. Within each cluster, we assess the spread of individual geometric parameters to understand which dimensions dominate variation. This provides insight into which physical attributes most influence flight-relevant dynamics and whether their tolerance ranges are wide enough to merit design attention or tight enough to be treated as fixed during manufacturing.

Each centroid model is thus treated as a valid design candidate and paired with a corresponding set of MPC weight parameters in the subsequent co-design optimization phase.

\subsection{Trajectory Generation}
\label{sec:trajectory_gen}

\begin{figure}[t]
      \centering
       \includegraphics[trim=350 0 350 0, clip, width=0.47\textwidth]{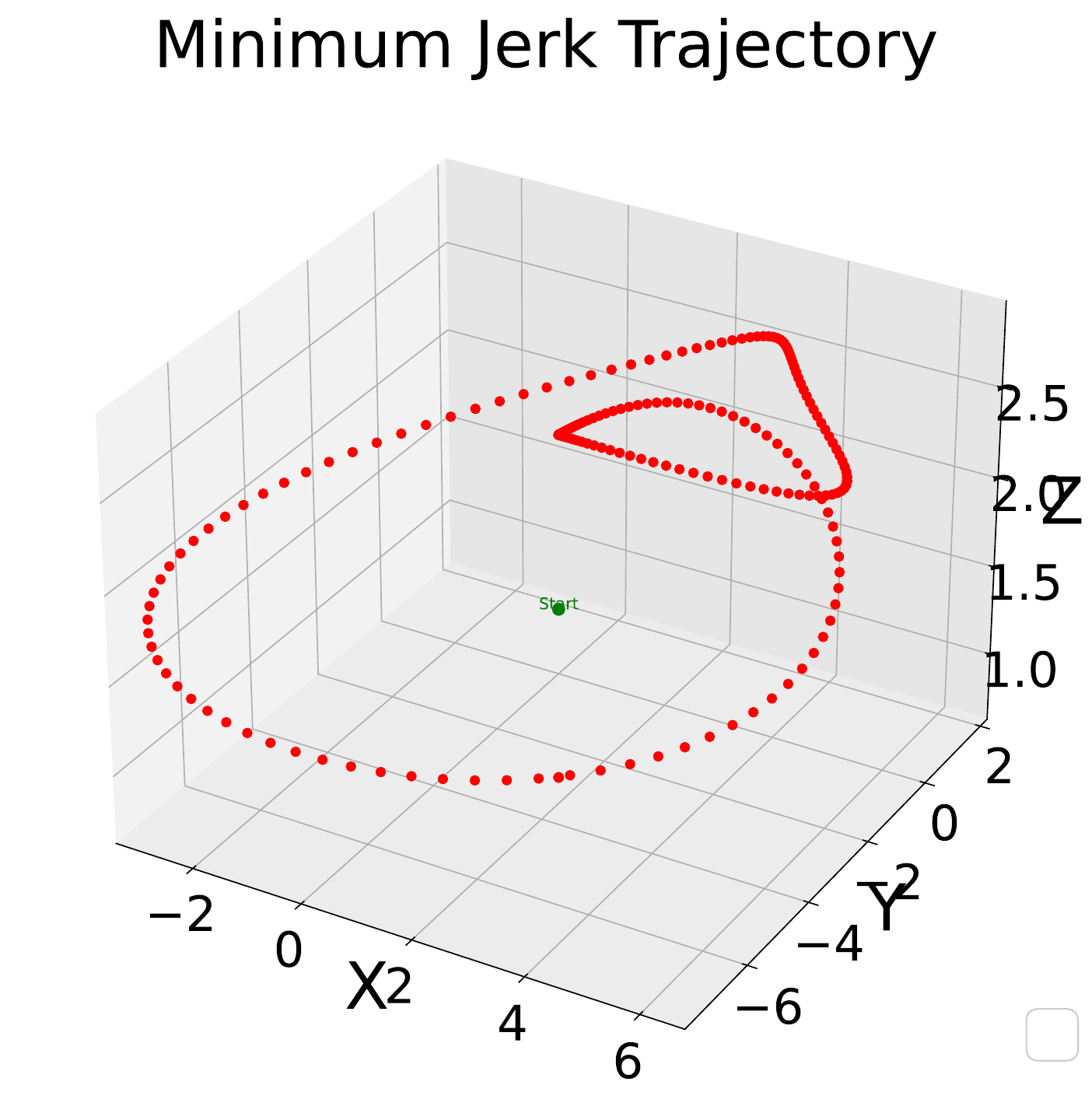}
      \caption{Minimum jerk trajectory interpolated over spatially placed waypoints.}
      \label{fig:gate_track}
   \end{figure}

The evaluation trajectory is defined as a sequence of spatial waypoints in 3D space, as shown in Figure ~\ref{fig:gate_track}. Each waypoint specifies a desired position the robot must traverse during flight. Although orientation information is included at each waypoint, it is used solely to shape the interpolation direction and is not enforced during control.

A smooth minimum-jerk trajectory is generated by interpolating between these waypoints, taking into account the local directional cues to ensure consistent path progression. The resulting trajectory provides continuous position and velocity references with smooth derivatives, supporting stable and dynamically feasible motion throughout the flight sequence.

The trajectory generation outputs:
\begin{itemize}
    \item Center of Mass (CoM) positions
    \item Corresponding velocity references
\end{itemize}

\begin{figure*}
    \centering
       \includegraphics[width=1\textwidth]{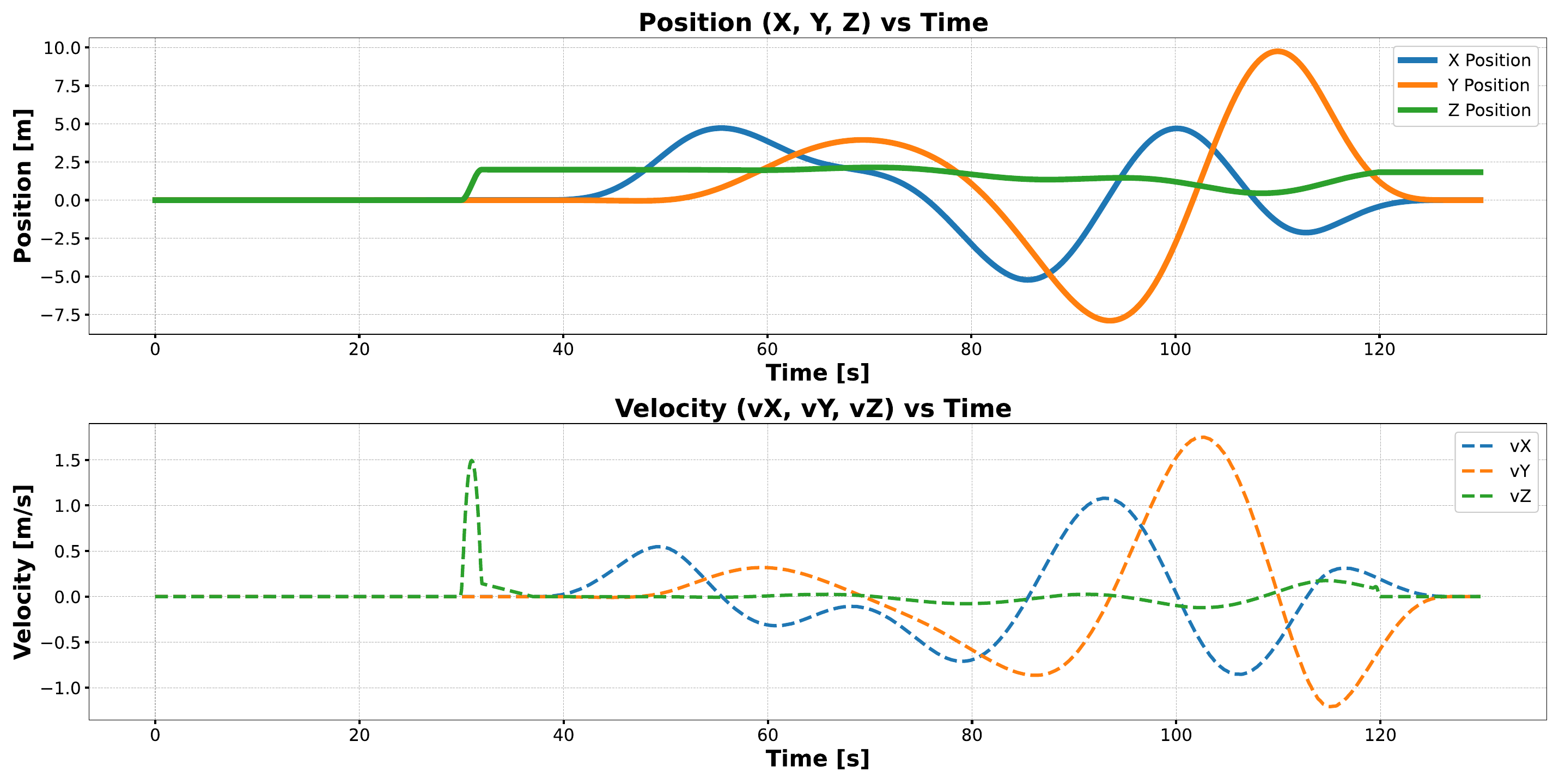}
      \caption{Reference position and velocities for CoM}
      \label{fig:pos_vel}
\end{figure*}

These references plotted in Figure \ref{fig:pos_vel} are used as inputs to the MPC controller. By keeping the trajectory fixed for all centroid models and weight combinations, the framework ensures consistency across evaluations and enables fair comparison of trajectory tracking and energy performance.
\subsection{Design–Control Optimization}
\label{sec:design_control_opt}

To identify suitable combinations of robot morphology and control parameters, we formulate a co-design optimization procedure based on a multi-objective genetic algorithm. The goal is to evaluate centroid models and associated MPC weights jointly, using trajectory-level performance metrics as evaluation criteria.

The search space is composed of:
\begin{itemize}
    \item A centroid index representing a valid, structurally feasible robot model.
    \item A vector of MPC cost weights used to tune the control effort across different state variables.
\end{itemize}

The optimization is performed using the NSGA-II algorithm, which ranks candidate solutions based on Pareto dominance across two primary objectives:

\subsubsection{Trajectory Tracking Error} This objective measures the deviation between the robot's center of mass (CoM) and the reference trajectory. For each simulation, the mean squared error (MSE) is computed separately for the \(x\), \(y\), and \(z\) directions. The total tracking error is expressed as:
    \[
    \text{MSE}_{\text{total}} = \left\|
    \begin{bmatrix}
    \text{MSE}_x \\
    \text{MSE}_y \\
    \text{MSE}_z
    \end{bmatrix}
    \right\|_2
    \]
    
\subsubsection{Energy Expenditure} To assess control effort, we compute the mechanical power exerted by the turbine thrusts over time. The instantaneous power includes contributions from both translational and rotational motion:
    \[
    P(t) = \mathbf{f}_{th}(t)^\top \mathbf{v}_{base}(t) + \boldsymbol{\tau}_{th}(t)^\top \boldsymbol{\omega}_{base}(t)
    \]
    The total energy expenditure is then integrated over the full trajectory duration with $\Delta t = 0.1$:
    \[
    E = \sum_{t=0}^{T} |P(t)| \cdot \Delta t
    \]
    By considering the absolute power, the cost accounts for both acceleration and deceleration phases—capturing thrust usage during braking and redirection. This encourages energy-efficient solutions that minimize unnecessary control effort across the full trajectory.

The co-design process operates on a fixed input population formed by ULH sampling over centroid indices and weight vectors. 
\begin{algorithm}[H]
\caption{Co-Design Optimization Framework}
\begin{algorithmic}[1]
\State Generate 5000 robot models with CAD-derived inertial and mesh properties.
\State Apply K-means clustering to obtain $N$ representative centroid models.
\State Generate a fixed minimum-jerk trajectory using spatially aligned waypoints.
\State Define a linearized MPC controller with 8 tunable cost function weights.
\State Create input population by ULH sampling over:
\begin{itemize}
    \item \textbf{Centroid model indices} 
    \item \textbf{MPC weight vectors}
\end{itemize}
\State Initialize NSGA-II with bi-objective cost:
\begin{itemize}
    \item Minimize CoM tracking error 
    \item Minimize absolute energy expenditure
   \end{itemize}
\For{each candidate}
    \State Simulate the trajectory with selected centroid and controller.
    \State Compute $\text{MSE}_{\text{total}}$ and Energy expenditure $E$.
    \If{QP fails or error exceeds threshold}
        \State Discard or penalize the candidate.
    \EndIf
\EndFor
\State Return the Pareto-optimal design–control combinations.
\end{algorithmic}
\end{algorithm}
For each candidate solution, a trajectory is simulated using the selected centroid model and MPC configuration.

To ensure meaningful evaluations, the following conditions are enforced:
\begin{itemize}
    \item Simulations must complete the full trajectory duration without MPC solver failure.
    \item Tracking error must remain below a threshold of \(2.5\,\mathrm{m}\).    
\end{itemize} 

Candidate solutions violating any of these constraints are penalized or excluded from the Pareto front. This ensures that the optimization is restricted to dynamically valid and physically realizable designs. \looseness=-1

%% file: Results.tex
This section presents the outcomes of the proposed co-design framework, focusing on how geometric clustering and control parameter tuning influence flight performance. The analysis aims to evaluate two key aspects: (i) the effect of clustering granularity on the diversity and quality of optimized solutions, and (ii) the role of individual designs in shaping the overall trade-off between trajectory tracking accuracy and energy expenditure. By comparing different clustering configurations and examining Pareto-optimal behaviors, we demonstrate the framework’s ability to identify designs that are both high-performing and structurally feasible for aerial tasks. \looseness=-1

The co-design framework is executed on a machine with an Intel(R) Core(TM) i7-10875H CPU and 32 GB RAM using  Windows 11 operating system. Geometric modeling and mass property extraction are performed in SolidWorks\textsuperscript{\textregistered}, with simulation-ready models exported in MuJoCo XML format for physics-based evaluation using the MuJoCo simulator \cite{todorov2012mujoco}. The momentum-based controller is implemented in C++ and interfaced via Python bindings. Optimization is managed through modeFRONTIER\textsuperscript{\textregistered}, with simulations run from a WSL2-based Conda environment. The code to reproduce the results is available online \cite{vantedducaddriven}. \looseness=-1

To investigate the effect of cluster granularity on co-design performance, the NSGA-II algorithm was executed using centroid sets of size 50, 100, and 250. Each centroid served as a representative design drawn from the full 5000-model space. The input parameter space for the optimization was constructed by combining centroid model indices with associated MPC gain parameters into unified candidate vectors. Initial populations were generated using ULH sampling to ensure diverse coverage across both morphological and control dimensions. The optimization targeted two objectives: minimizing trajectory tracking error and energy consumption over a predefined minimum-jerk trajectory.

Population sizes and generation counts were scaled across cluster resolutions to maintain comparable total evaluations as seen in Table \ref{tab:nsga_settings}. Crossover probability was set to $\mathit{90\%}$ to encourage exploration, and mutation probability to $\mathit{11\%}$, corresponding to one expected mutation in the nine-dimensional input space. Elitism and NSGA-II’s sorting and crowding distance mechanisms guided the selection process. This configuration allowed evaluation of how clustering affects the solution quality and parameter diversity in the co-design process.\looseness=-1

\begin{table}[h]
    \scriptsize
    \centering
    \caption{NSGA-II Settings Across Clustering Resolutions}
    \label{tab:nsga_settings}
    \resizebox{\columnwidth}{!}{%
    \begin{tabular}{c|c|c|c}
        \toprule
        \textbf{Clusters} & \textbf{ Initial Pop. Size} & \textbf{Generations}  & \textbf{Time (hrs)} \\
        \midrule
        \rowcolor{gray!10}
        50 & 25 & 60 & 15 \\
        100 & 40 & 60  & 26 \\
        \rowcolor{gray!10}
        250 & 100 & 60  & 75 \\
        \bottomrule
    \end{tabular}
    }
\end{table}
\begin{figure}[h]
    \centering
    \includegraphics[width=1\linewidth]{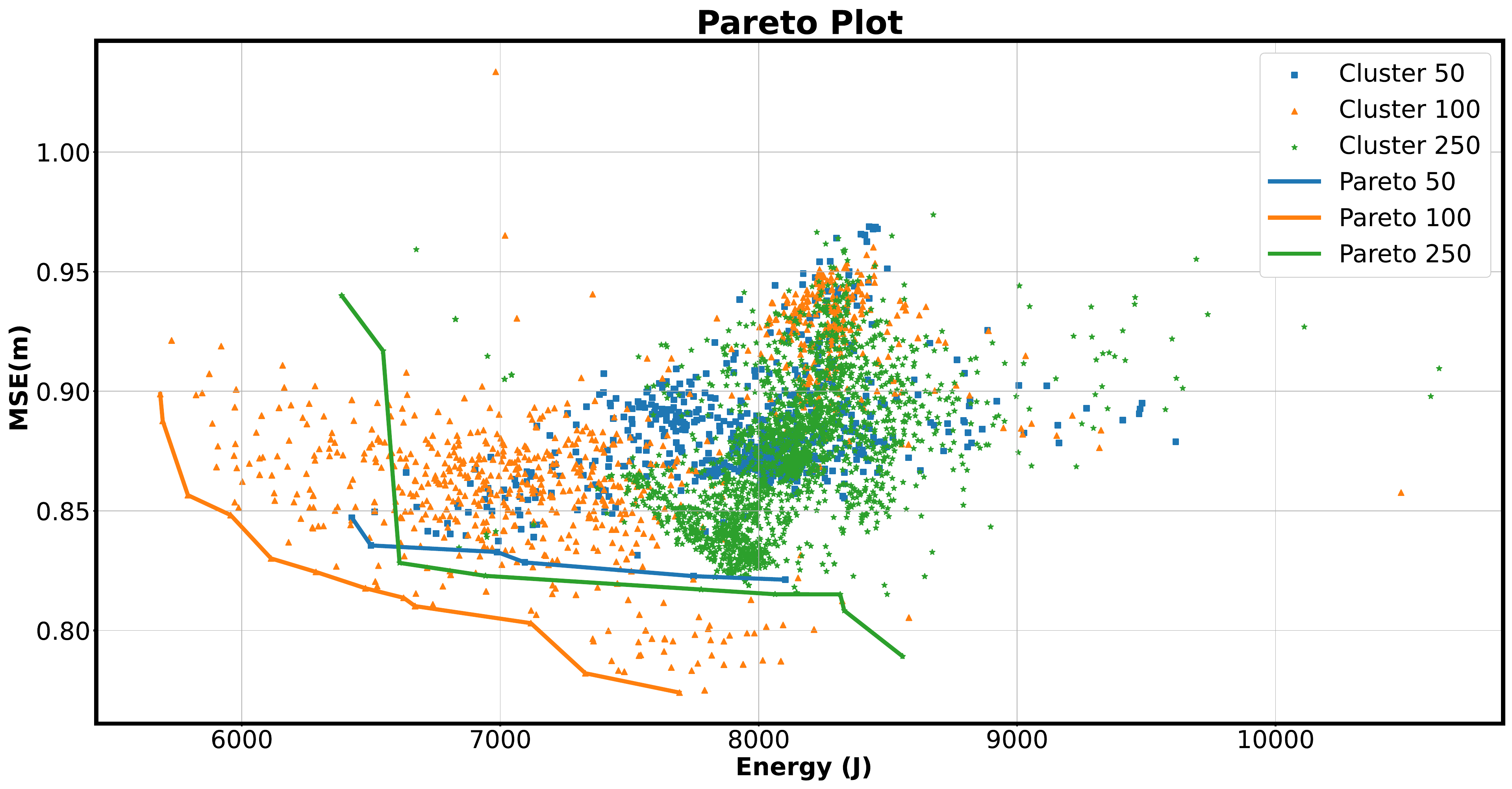}
    \caption{Pareto front showing trade-off between tracking error and energy consumption. }
    \label{fig:Pareto}
\end{figure}

Figure~\ref{fig:Pareto} shows the scatter of candidate solutions and corresponding Pareto fronts for the 50, 100, and 250 cluster cases, evaluated across the two objectives: trajectory tracking error and energy expenditure. While all three cluster resolutions yield non-dominated solutions, the 100-cluster case exhibits a consistently lower Pareto front, indicating improved trade-offs between performance and energy. In contrast, the fronts for 50 and 250 clusters appear poorly converged, with wider spread and higher objective values. This may be attributed to the granularity of clustering: the 50-cluster case likely lacks sufficient morphological diversity, limiting optimization potential, conversely, the 250-cluster case increases search complexity, diluting computational effort across a larger number of centroids. The results suggest that intermediate clustering resolutions, such as 100, offer a more effective balance between design diversity and optimization traceability. 

\begin{figure*}     
    \begin{subfigure}[t]{0.48\textwidth}
        \centering
        \includegraphics[width=\linewidth]{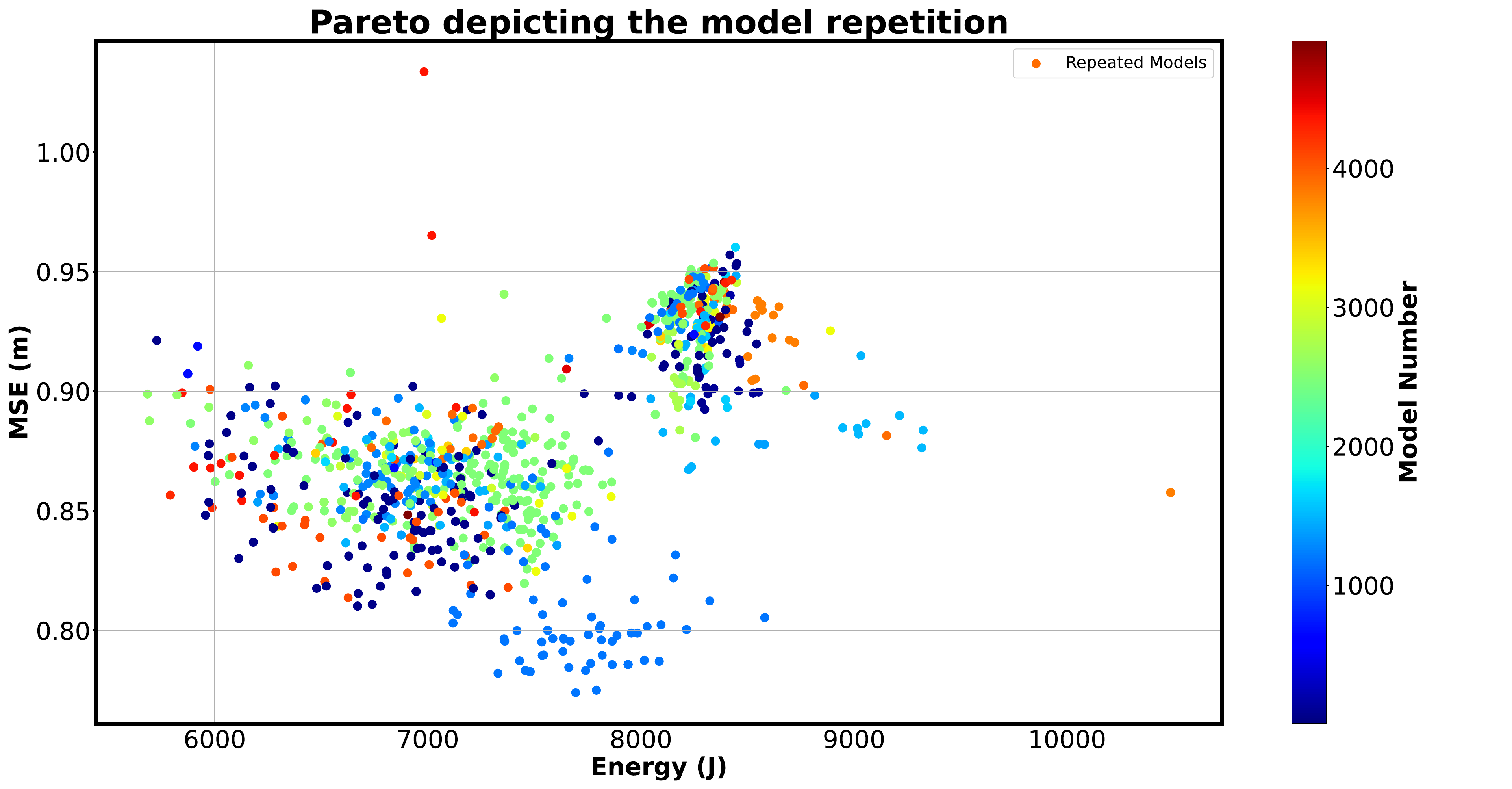}
        \caption{}
        \label{fig:ParetoRepetition}
    \end{subfigure}
    \hfill
    \begin{subfigure}[t]{0.48\textwidth}
        \centering
        \includegraphics[width=\linewidth]{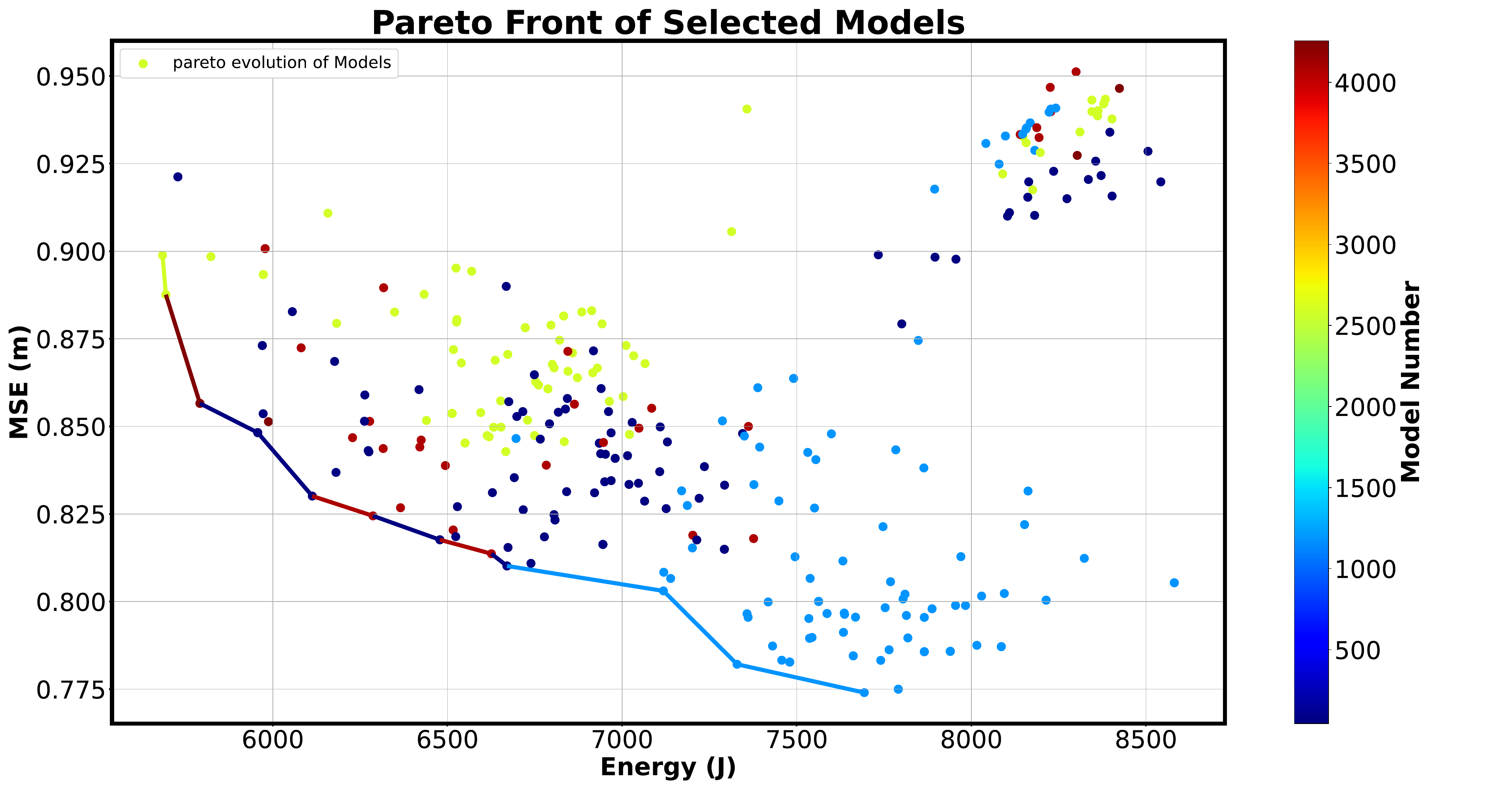}
        \caption{}
        \label{fig:pareto_evolution}
    \end{subfigure}
    \caption{
    Multi-objective optimization results for the 100-cluster configuration. 
    (\subref{fig:ParetoRepetition}) Distribution of candidate solutions, color-coded by centroid model index. Repetition and spatial clustering indicate which morphologies consistently yield high-performing outcomes across varied MPC gain settings.
    (\subref{fig:pareto_evolution}) Evolution of selected centroid models contributing to the final Pareto front. Each point represents a design–controller pair, and connected segments highlight how tuning control parameters shifts performance in the energy–error space, forming the structure of the Pareto frontier.
    }
    \label{fig:pareto_combined}
\end{figure*}

To further analyze how individual designs influence the optimization outcomes, Figure~\ref{fig:ParetoRepetition} shows the distribution of all candidate solutions from the 100-cluster configuration, color-coded by centroid model index. Each point represents a unique design–controller combination evaluated during the NSGA-II optimization. The repetition and spatial grouping of certain models across the objective space indicate that specific morphologies consistently yield high-performing solutions under various control gain configurations.

\begin{table*}[h]
    \centering
    \caption{Geometric parameters of selected centroid models}
    \label{tab:centroid_models}
    \setlength{\tabcolsep}{5pt} 
    \renewcommand{\arraystretch}{1.2} 
    \begin{tabular}{c|c|c|c|c|c|c|c|c}
        \toprule
        \textbf{Model ID} & \textbf{Angle [\qty{}\degree]} & \textbf{Offset [mm]} & \textbf{Height [mm]} & \textbf{Forearm [mm]} & \textbf{Shoulder [mm]} & \textbf{Hip [mm]} & \textbf{Ankle [mm]} & \textbf{Foot [mm]} \\
        \midrule
        \rowcolor{blue!60}
        44 (Blue) & 8 & 5 & 0 & 35 & 5 & 0 & 50 & 90 \\    
        \rowcolor{cyan!65 }
        1188 (Cyan) & 14 & 30 & 0 & 15 & 40 & 10 & 10 & 30 \\
        \rowcolor{green!60}
        2586 (Green) & 4 & 30 & 10 & 35 & 30 & 40 & 25 & 90 \\
        \rowcolor{red!50}
        4082 (Red) & 12 & 25 & 5 & 15 & 15 & 10 & 5 & 40 \\
        \rowcolor{red!60}
        4256 (dark Red) & 6 & 30 & 25 & 40 & 15 & 50 & 40 & 70 \\
        
        \bottomrule
    \end{tabular}
\end{table*}
To isolate the most influential models, we extracted the non-dominated set of solutions that make up the final Pareto front from the 100-cluster run. The corresponding centroid model indices were identified, and those appearing repeatedly or occupying key regions of the front were selected for further analysis. Figure~\ref{fig:pareto_evolution} highlights the selected models—specifically indices \textbf{44}, \textbf{1188}, \textbf{2586}, \textbf{4082}, and \textbf{4256}—by showing their distribution in the energy–error space. Each point corresponds to a specific set of MPC gains applied to one of these designs. For models with multiple Pareto-optimal solutions, the points are connected with line segments of the same color to trace their spread across the front. This reveals how adjusting the controller weights for a fixed morphology results in different trade-offs between tracking error and energy consumption.\looseness=-2

Models \textbf{1188} and \textbf{2586} dominate the extremes of the Pareto front. Model 1188 achieves low tracking error at high energy cost, while model 2586 minimizes energy consumption at the expense of accuracy. The consistent color along their paths indicates that these designs specialize in optimizing a single objective. Conversely, models \textbf{44}, \textbf{4082}, and \textbf{4256} appear in the central "knee" region of the trade-off curve. The presence of multiple colors and overlapping contributions in this region shows that these designs are more adaptable, offering well-balanced solutions between accuracy and efficiency through gain tuning.

The specific geometric parameters corresponding to these five centroid models are listed in Table~\ref{tab:centroid_models}. While no direct or consistent correlation was found between specific geometric features and performance outcomes, the repeated appearance of certain models across key regions of the Pareto front highlights their importance in the co-design process. This analysis confirms that multiple high-performing solutions can arise from the same design through control tuning and that some morphologies are more frequently associated with favorable trade-offs.

To summarize, this section demonstrates how the proposed co-design framework effectively identifies robot designs that offer favorable trade-offs between trajectory tracking accuracy and energy consumption. By clustering a large set of CAD-derived models and optimizing control parameters using NSGA-II, we isolated a small subset of centroid morphologies that consistently contribute to the Pareto front. The analysis confirms that certain designs are not only structurally feasible but also more adaptable under gain tuning, making them strong candidates for real-world deployment. This validates the joint design–control optimization strategy as a practical method for guiding the selection of flight-capable aerial humanoid configurations.

%% file: Conclusions.tex
This paper presented a CAD-driven co-design framework for jet-powered aerial humanoid robots, combining morphological variation with control parameter tuning. Starting from the iRonCub-Mk3 platform, a large set of geometrically valid designs was generated through parametric CAD modeling and DoE sampling. Each design was evaluated in simulation using a linearized MPC controller tracking a predefined trajectory. A multi-objective optimization using NSGA-II was employed to identify combinations of design and control parameters that minimize tracking error and energy consumption.

To manage the evaluation cost, K-means clustering was used to reduce the design space to a set of representative centroids. The results showed that meaningful trade-offs could be achieved and that several designs consistently produced good performance when paired with appropriate control gains.

The current implementation is limited by evaluation time and the use of fixed NSGA-II hyperparameters, such as crossover and mutation rates. While higher-resolution clustering may yield a more diverse set of optimal solutions, it would require parallelized evaluation to remain computationally practical. In future work, we aim to extend the framework to support different geometric variations and apply it to a wider range of locomotion and manipulation tasks. This will help assess the generalizability of CAD-based co-design across various robot morphologies and mission profiles, enabling more flexible and efficient design workflows.\looseness=-1